\documentclass{article}

\usepackage{arxiv}
\usepackage{times}
\usepackage{epsfig}
\usepackage{graphicx}
\usepackage{amsmath}
\usepackage{amssymb}

\usepackage{array}
\usepackage{glossaries}
\usepackage{subfig}
\usepackage{multicol}
\usepackage{threeparttable}
\usepackage{longtable}
\usepackage{fancyvrb}
\usepackage{changepage}
\usepackage[stable]{footmisc}
\usepackage[T1]{fontenc}
\usepackage{multirow}
\usepackage{mathtools}
\usepackage{siunitx}

\usepackage{color}

\usepackage[linesnumbered,ruled]{algorithm2e}
\SetAlgorithmName{Procedure}{procedure}{List of Procedures}

\usepackage{booktabs}

\DeclareMathOperator*{\argmax}{argmax}

\newcommand{\etal}{\textit{et al}.}

\usepackage[pagebackref=true,breaklinks=true,letterpaper=true,colorlinks,bookmarks=false]{hyperref}

\begin{document}

\title{Quantization in Relative Gradient Angle Domain For Building Polygon Estimation}

\author{Yuhao Chen \qquad Yifan Wu \qquad Linlin Xu \qquad Alexander Wong\\
\and
Vision and Image Processing Lab (VIP) \\
Systems Design Engineering Department \\
University of Waterloo \\
Waterloo, Ontario, Canada
}

\maketitle

\begin{abstract}
Building footprint extraction in remote sensing data benefits many important applications, such as urban planning and population estimation.
Recently, rapid development of Convolutional Neural Networks (CNNs) and open-sourced high resolution satellite building image datasets have pushed the performance boundary further for automated building extractions.
However, CNN approaches often generate imprecise building morphologies including noisy edges and round corners.
In this paper, we leverage the performance of CNNs, and propose a module that uses prior knowledge of building corners to create angular and concise building polygons from CNN segmentation outputs.
We describe a new transform, Relative Gradient Angle Transform (RGA Transform) that converts object contours from time vs. space to time vs. angle.
We propose a new shape descriptor, Boundary Orientation Relation Set (BORS), to describe angle relationship between edges in RGA domain, such as orthogonality and parallelism.
Finally, we develop an energy minimization framework that makes use of the angle relationship in BORS to straighten edges and reconstruct sharp corners, and the resulting corners create a polygon.
Experimental results demonstrate that our method refines CNN output from a rounded approximation to a more clear-cut angular shape of the building footprint.

\end{abstract}

\section{Introduction}

Building footprint extraction from remote sensing data has many important applications, such as urban planning, tax estimation, population estimation, and energy demand estimation.
Manual labeling building footprints can be labor-intensive and time-consuming.
Therefore, automated building extraction methods are crucial for making the above applications cost-effective.
Recently, open-sourced satellite building image datasets and rapid development of Convolutional Neural Networks (CNNs) have pushed the performance boundary further for automated building extractions.
Among the datasets, high resolution satellite image datasets provide astonishing building details and precise building footprints, including Spacenet~\cite{spacenet}, Inria Building dataset~\cite{inria_dataset}, and Deep Globe~\cite{deepglobe}. 
In our study, we focus on Inria Building dataset for its relatively accurate labels.

\begin{figure}[]

	\centering
	\subfloat[]{\label{fig:method:ori}{\epsfig{figure=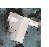,width = 0.21\textwidth}}}
	\qquad 
	\subfloat[]{\label{fig:method:res}{\epsfig{figure=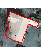,width = 0.21\textwidth}}}
	\qquad 
	\subfloat[]{\label{fig:method:prob}{\epsfig{figure=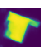,width = 0.21\textwidth}}}
	\qquad 
	\subfloat[]{\label{fig:method:polygon}{\epsfig{figure=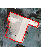,width = 0.21\textwidth}}}
	\caption{a) An example building.
			 b) Segmentation output of a CNN network. Contour of the output is marked in red.
			 c) Input of our module, the building probability map $P_n$ produced by a CNN segmentation network.
	         d) Polygon extracted by our method.}
	\label{fig:method:example}
\end{figure}

Convolutional Neural Networks (CNNs) such as Deeplab, UNet, and Fully Convolutional Networks have delivered promising results for semantic segmentation~\cite{deeplab, unet, fcn}.
Based on the above semantic segmentation networks, numerous methods have been developed for building extraction in specific~\cite{yuan2017learning, bischke2019multi, ji2018fully, yang2018building, iglovikov2018ternausnetv2, li2018semantic, golovanov2018building, bittner2018building}.
Typical building labels for CNN training include more content pixels than boundary pixels.
This imbalance often causes CNN to produce inaccurate building edges. 
Some studies utilize distance transform as additional information to enhance building boundaries.
However, CNN approaches focus on textures derived from convolutional filters, and do not consider the spatial continuity and smoothness on object's boundaries.
As a result, building edges may not be straight even with the enhancement from distance transform.
To address these issues, several studies developed end-to-end networks for building polygon estimation~
\cite{marcos2018learning, cheng2019darnet, zhang2019conv, nauata2019vectorizing}. 
There are two approaches, active contour approach and edge assemble approach.
Active contour \cite{act_contour} approach considers both accuracy and smoothness of edges in its loss function. 
Yet, its smoothness term discourages sharp corners which most building has.
The edge assemble approach first use CNN to detect building edges and corners, then assembles them into polygons.
This approach builds on the premise that all building edges are detected.
Detecting building edges can be difficult when there's an occlusion.
A missing edge can result in a collapse of the entire polygon. 
In summary, existing methods do not perform well in producing smooth edges, sharp corners, and handling occlusions at the same time.

\begin{figure*}[t]
	\centering
	\centerline{{\epsfig{figure=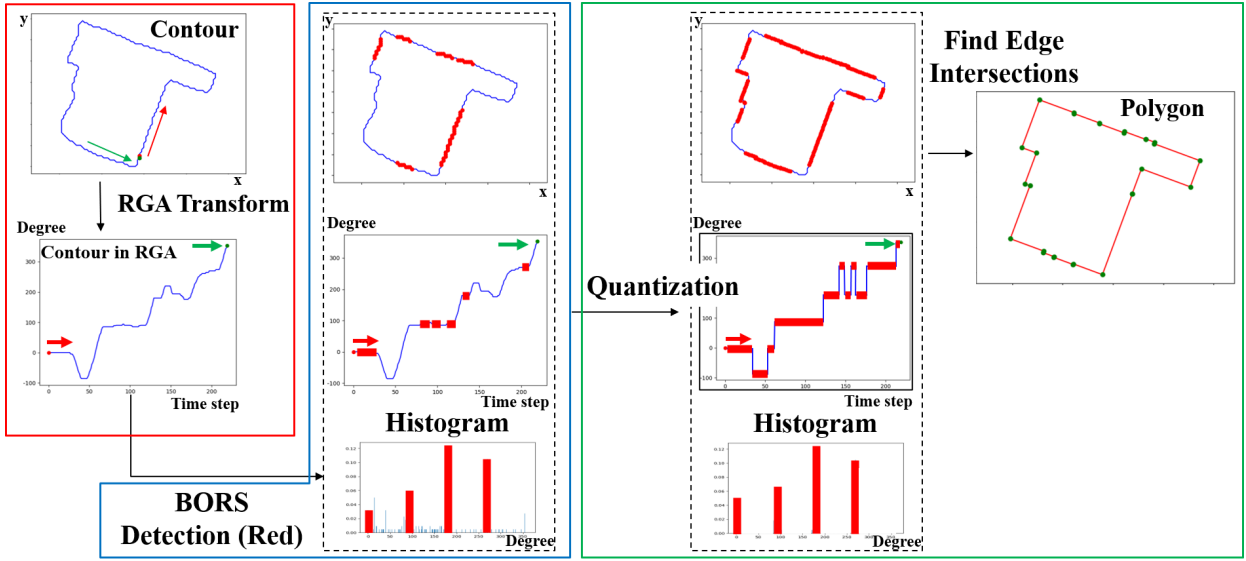,width = 1\textwidth}}}
	\caption{Block Diagram of our method. Red box: Relative Gradient Angle Transform (RGA Transform) converts object contours from time vs. space to time vs. angle. (Section \ref{sec:RGAT})
	Blue box: Detecting a shape described by Boundary Orientation Relation Set (BORS). Red points and bars correspond to the value associated with the detected BORS in various domains.  (Section \ref{sec:bors})
	Green box: Quantizing angles in RGA domain using an energy minimization framework that makes use of the angle relationship in BORS to straighten edges and reconstruct building corners. The resulting corners create a polygon. (Section \ref{sec:quantization})}
	\label{fig:block_diagram}
\end{figure*}

In this paper, we leverage the performance of CNNs, and propose a module that uses prior knowledge of building corners to create angular and concise building polygons from CNN segmentation outputs.
We describe a new transform, Relative Gradient Angle Transform (RGA Transform) that converts object contours from time vs. space to time vs. angle.
We propose a new shape descriptor, Boundary Orientation Relation Set (BORS), to describe angle relationship between edges in RGA domain, such as orthogonality and parallelism.
Finally, we develop an energy minimization framework that makes use of the angle relationship in BORS to straighten edges and reconstruct sharp corners, and the resulting corners create a polygon.
Experimental results demonstrate that our method refines CNN output from a rounded approximation to a more clear-cut angular shape of the building footprint.
Figure \ref{fig:block_diagram} shows the block diagram of our method.

\section{Related Work}

\textbf{Semantic Segmentation Using CNNs:} Pixel-level semantic segmentation is a key task and an active topic in computer vision. 
The emergence of CNNs has enabled great advancement in this field. 
Fully Convolutional Networks (FCNs) \cite{fcn} introduces deconvolution as upsampling operations to replace fully connected layers in classification models. 
U-Net \cite{unet} uses multiple upsampling layers and skip connections to improve upon FCNs. 
SegNet \cite{badrinarayanan2017segnet} adopts encoder-decoder architecture to produce dense feature maps. 
In order to integrate information from various spatial scales, PSPNet \cite{zhao2017pyramid} features a pyramid pooling module to distinguish patterns with different scales.
FPN \cite{feature_pyramid_network} employs lateral connection merging feature maps from the bottom-up pathway and the top-down pathway. 
DeepLab \cite{deeplab} makes use of both dilated convolutions \cite{yu2015multi} and fully connected Conditional Random Fields (CRFs) \cite{krahenbuhl2011efficient} as post-processing step to incorporate both local and global information. 

\textbf{Building Footprint Extraction:} The automation of building footprint extraction is an important problem in remote sensing. 
Early methods focus on utilizing height or 3D geometry information.
Studies in \cite{wang2006bayesian, zhang2006automatic} propose approaches to extract building footprints from LIDAR data. 
In \cite{wang2006bayesian}, a preliminary building footprint is estimated by the shortest path \cite{algorithm_book}, and the footprint is refined by maximizing a posterior probability. 
In \cite{zhang2006automatic}, the contour of the building is refined by smaller operations including split, intersect, merge, and remove. 
These operations involve a lot of thresholds which make the method non-robust.
In \cite{bredif2013extracting, tournaire2010efficient}, Bredif~\etal use height information in digital surface models (DSMs) and digital elevation models (DEMs) to tackle the problem with energy functions and assumption of rectangular building shape.

Recent CNNs show promising improvement in semantic segmentation performance which also allows great progress been made to high-resolution aerial and satellite imagery analysis. 
U-Net is a popular baseline.
In \cite{ji2018fully, iglovikov2018ternausnetv2}, Ji~\etal and Iglovikov~\etal develop modifications to U-Net targeting building extraction.
In \cite{li2018semantic}, Li~\etal apply threshold and post-processing to U-Net predictions. 
Other works rely on additional input information for further improvements. 
In \cite{bittner2018building}, Bittner~\etal fuse RGB, panchromatic, and normalized DSM data as CNN inputs to produce better building segmentation. 
In \cite{yuan2017learning}, Yuan~\etal introduce the signed distance function of building boundaries to improve CNN output representation. 
In \cite{yang2018building, bischke2019multi}, Yang~\etal and Bischke~\etal make use of additional distance-transformed building labels as input to CNNs exploring better preservation of boundaries. 
In \cite{golovanov2018building}, a composite loss function and weighted building labels are used for the same purpose. 

\textbf{Polygon-based Building Boundary Delineation:} Building boundary is considered the most important feature of a building footprint because it defines the shape and location of the building. 
While pixel-based CNN approaches produce results with high recall, they are not good at building boundary delineation. 
Accurate predictions with sharp corners and straight building edges are hard to achieve. 
To solve this challenge, Marcos~\etal \cite{marcos2018learning} propose deep structured active contours (DSAC) framework which combines CNNs and active contour model (ACM) \cite{act_contour} to produce a polygon-based output model that is trainable end-to-end. 
Cheng~\etal \cite{cheng2019darnet} improve it by representing contour points in polar coordinates as active rays.

Although the approaches above improve mask coverage compared to pure CNN-based segmentation, blob-like contours that do not closely assemble building boundaries still exist.
In \cite{zhang2019conv}, Zhang~\etal employ Graph Neural Networks to reconstruct building planar graph from high-resolution satellite imagery. 
In \cite{nauata2019vectorizing}, Nauata~\etal detect corners, edges and regions using CNN, and assemble them using integer programming. 
Those methods generate polygonal building output, but the usability of them is limited by the requirement of extra building corner and edge annotations for planar graph reconstruction learning.

\section{Assumptions and Preprocessing}

In this paper, we assume all shapes are closed. 
To simplify the notation, we use a circular condition for index calculations.
For example, given a shape with length $l$ and an index $i$, index $(i \bmod l)$ will be used for any index calculation.
In addition, all angles are described in degree.

Let $P_n$ to be the probability map of network output. 
Figure \ref{fig:method:prob} shows an example probability map for a building.
A building pixel in $P_n$ has a value closer to 1, and non-building pixels have a value closer to 0.
Its corresponding building segmentation mask $m_s$ is obtained by thresholding $P_n$ with probability 0.5:
\begin{equation}
    m_s(x, y) = \begin{cases}
        1 & P_n(x, y) > 0.5\\
        0 & otherwise , 
    \end{cases}
\end{equation}
where $(x, y)$ is a coordinate in the mask.

We extract individual buildings in $m_s$ using Connected Components Analysis \cite{connected_components}.
Let $m_c$ to be the segmentation mask of a connected component.

\section{Relative Gradient Angle Transform}
\label{sec:RGAT}
\begin{figure}[]
	\centering
	\subfloat[]{\label{fig:method:gradient_angle}{\epsfig{figure=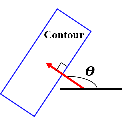,width = 0.18\textwidth}}}
	\qquad
	\subfloat[]{\label{fig:method:contour}{\epsfig{figure=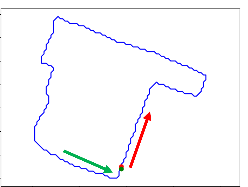,width = 0.2\textwidth}}}
		\\
	\subfloat[]{\label{fig:method:rga}{\epsfig{figure=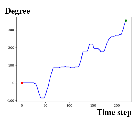,width = 0.36\textwidth}}}
	\caption{a) Gradient angle $\theta$ of a contour.
			 b) Contour of a building. Red: contour start. Green: contour end.
	         c) Relative Gradient Angles along the contour.}
\end{figure}

Gradient angle is an important feature along an object's boundary.
Figure \ref{fig:method:gradient_angle} shows an example of a gradient angle.
We want to represent an object's contour by time vs. angle instead of typical time vs. space.
Angle ranges involved in trigonometry computations are limited to $[0, 360)$ instead of $(-\inf, \inf)$.
When converting contour points from position to gradient angle, the bounded angle range creates two problems.
First, similar angles may have large numerical differences.
For example, 0\si{\degree} and 359\si{\degree} have a numerical difference of 359 degrees, but the actual difference between the two is 1 degree.
This makes the gradient angle signal spiky and difficult to analyze.
The second problem is the following.
Ideally, gradient angle goes from 0\si{\degree} to 360\si{\degree} along an object's contour in one cycle.
However, when computing gradient angles using $\arctan$ function, every angle gets mapped to $[-90, 90)$ range. 
To determine whether we need to add 360 degrees (produces the same angle with different numerical value) or 180 degrees (produces opposite angle with the same $\tan$ value) to each angle is difficult. 
Therefore, direct conversion is problematic for contours from time vs. space domain to time vs. angle domain.

Instead of computing gradient angles for individual contour points, we find the relative difference between adjacent gradient angles is more representative for describing the shape of a contour.
If we define the signed gradient angle to be positive pointing inwards of the object's mask, we observe that positive angle difference indicates convexity and negative angle difference indicates concavity of object's shape at the angle difference location.
Therefore, we propose Relative Gradient Angle (RGA) Transform to convert contour signals into gradient angle signals along an object's contour.

We apply contour extraction on the mask $m_c$ using the method described in \cite{find_contour}.
We obtain an initial contour signal $C_0=\{c_0, c_1, ..., c_{n-1}\}$, where $n$ is the number of points and $c_i=(x_{c_i}, y_{c_i})$ is the $i$th contour point with image coordinate $(x_{c_i}, y_{c_i})$.
Computing tangent angles directly on the initial contour set can produce noisy results due to spatial aliasing.
Therefore, we smooth contour using moving average filter with a window size of $W_a$.
We obtain a smoothed contour $C_s$.
Figure \ref{fig:method:contour} shows the smoothed contour of a building.
We define the middle point of each adjacent contour point pair to be $q_i$, where $q_i = (\frac{x_{c_i}+x_{c_{i+1}}}{2}, \frac{y_{c_i}+y_{c_{i+1}}}{2}), c_i \in C_s$ and $c_{i+1} \in C_s$.
The tangent angle $t_i$ at the middle point $q_i$ is 
\begin{equation}
    t_i = \arctan{(\frac{y_{c_{i+1}}-y_{c_{i}}}{x_{c_{i+1}}-x_{c_{i}}})}
\end{equation}
Now, we compute gradient angles at object's contour.
We define the gradient angle to be the orthogonal angle $o_i$ pointing inward of the object's mask at $q_i$:
\begin{equation}
    o_i = \begin{cases}
        t_i & \text{$t_i$ points inward of the mask}\\
        t_i + 180 & \text{$t_i$ points outward of the mask}
    \end{cases}
\end{equation}

Let the contour signal in RGA domain to be $\Theta_0 = \{\theta_0, \theta_1, ..., \theta_{n-1}\}$.
We call each contour point in RGA domain a contour angle, and the contour signal in RGA domain a contour angle signal.
We define the $i$th contour angle $\theta_{i}$ as the following
\begin{equation}
    \theta_{i} = \begin{cases}
    			0 & i = 0\\
    			\theta_{i-1} + d(o_{i}, o_{i-1}) & otherwise,
    			\end{cases}
\end{equation}
where $d(*)$ is a function that subtracts angle $o_{i-1}$ from $o_{i}$ and returns the difference in $[-90, 90)$ range.
The set $\{\theta_0,\theta_1,\cdots,\theta_{n-1},\theta_n\}$ always starts with $0\si{\degree}$ and ends with $360\si{\degree}$, as $\theta_n$ can be equal to either 0$\si{\degree}$ or $360\si{\degree}$.

\section{Boundary Orientation Relation Set}
\label{sec:bors}
For each type of object, we assume there is an angle structure in the contour angle signal.
In other words, angles with a fixed relation co-occur in the contour angle signal. 
For example, in building applications, relationships between angles can be orthogonal or parallel.
We name the set of angle relationships a Boundary Orientation Relation Set (BORS).
We define a BOR set as $S = \{s_0, s_1 ..., s_{n_S-1}\}$, where $n_S$ is the number of relations.
Let $\psi=\{S_1, S_2, ..., S_{n_{\psi}}\}$ to be a bank of $n_{\psi}$ relation sets.
Each BOR set represents a type of object's shape.
For example, in this paper, we predefine the BOR set for buildings as $S = \{90, 180, 270\}$.

To find the best BOR set describing a shape, we first use median filter \cite{median_filter} with a window size of $W_m$ to filter the noise in the contour angle signal $\Theta_0$.
Unlike other filters, median filter preserves angle values that are in the contour angle signal.
We obtain the processed contour angle signal $\Theta$.
We then compute the contour angle distribution denote as $P_{\Theta}(\alpha)$ with variable $\alpha \in [0, 360) $ degrees.
Let $\alpha_0$ to be an initial angle.
Then the contour angles associated with the BOR set is $A_{S} = \{ \alpha_0, \alpha_0+s_0, \alpha_0+s_1, ..., \alpha_0+s_{n_{S}-1}\}$.
We call $A_{S}$ a structure angle set, and each angle in $A_S$ is a structure angle.

\begin{figure}[]
	\centering
	\subfloat[]{\label{fig:method:hist}{\epsfig{figure=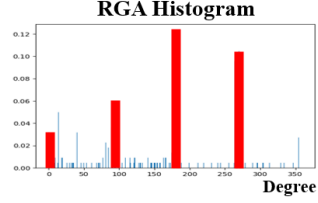,width = 0.26\textwidth}}}
	\\
	\subfloat[]{\label{fig:method:structure}{\epsfig{figure=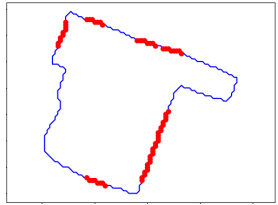,width = 0.22\textwidth}}}
	\caption{a) Contour angle distribution for the contour angle signal. Red indicates probabilities related to the BOR set $\{90, 180, 270\}$.
	         b) Structure edges marked in red.}
\end{figure}

To find the best relation set, we first define the probability of the contour angle signal $\Theta$ containing a relation described by the BOR set $S$ with initial angle $\alpha$ to be 
\begin{equation}
    P_{{\Theta}, S}(\alpha) = \sum^{n_{S}-1}_{l=0}{P_{\Theta}(\alpha+s_l)}
\end{equation}
Figure \ref{fig:method:hist} shows the histogram of a contour angle signal.

The initial angle $\hat{\alpha}_S$ for BOR set $S$ can be estimated using Maximum Likelihood Estimation (MLE):
\begin{equation}
    \hat{\alpha}_S = \argmax_{\alpha \in [0,360)}{P_{{\Theta}, S}(\alpha)}
\end{equation}

The optimal BOR set $S$ can be estimated using MLE as well:
\begin{equation}
	\hat{S} = \argmax_{S \in \psi}{P_{{\Theta}, S}(\hat{\alpha}_{S})}
\end{equation}

Figure \ref{fig:method:structure} shows the contour points associated with the optimal BOR set.

\section{Shape Refinement and Reconstruction by Quantization Based Energy Minimization}
\label{sec:quantization}

\begin{figure}[]
	\centering
	\centerline{{\epsfig{figure=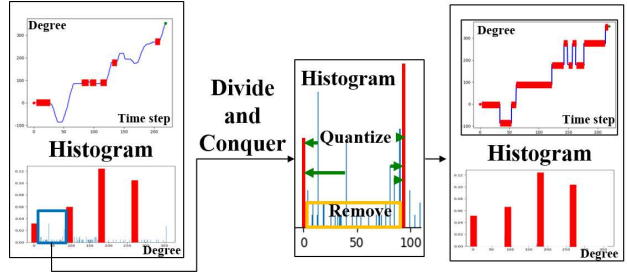,width = 0.5\textwidth}}}
	\caption{Block Diagram of the energy minimization framework.
	Given a contour angle signal and a Boundary Orientation Relation Set, we divide contour angle signal and minimize individual transitions (marked in blue box) between structure angles (marked in red). 
	The minimization includes removing angles with low occurrence and quantizing angles with high occurrence. 
	The images on the right shows the result after energy minimization.}
	\label{fig:block_diagram_quan}
\end{figure}

Ideally, all angles in a contour angle signal is a structure angle.
So, $P_{{\Theta}, S}({\alpha}_S) = 1$, meaning that the contour angle signal can be quantized to the structure angle set under the relation set $S$ without any loss. 
It also indicates the transitions between structure angles are abrupt.
In reality, network outputs round corners where the transitions between structure angles are more gradual.
These transitions contribute to the non-structure angle probabilities in the histogram. 
We replace round corners with sharp corners by quantizing contour angles to their nearest structure angle.
However, quantizing noise to its nearest angle can amplify the uncertainty of noise and create an oscillation effect in the contour angle signal, which results in a "step" effect in the contour signal.
Therefore, in our energy minimization framework, we identify and remove noises before applying quantization to contour signals. 
Figure \ref{fig:block_diagram_quan} shows the block diagram for the energy minimization framework.
 
Let $S$ to be the estimated BOR set, and let $A_{S}$ to be the structure angle set.
We adjust the contour angle signal to maximize the probability of the relation set $S$:
\begin{equation}
    g_S(\Theta) = {P_{\Theta, S}(\alpha_S)} \, .
\end{equation}
Equivalently, we minimize the transition between structure angles, and propose the energy function below:
\begin{equation}
    f_S(\Theta) = {(1-g_S(\Theta))} = {\sum_{\beta \notin A_{S}}{P_{\Theta}(\beta)}}
\end{equation}

\begin{figure}[]
	\centering
	\subfloat[]{\label{fig:method:rga1_pts}{\epsfig{figure=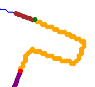,width = 0.18\textwidth}}}
	\quad
	\subfloat[]{\label{fig:method:edge_new_line}{\epsfig{figure=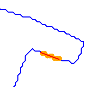,width = 0.18\textwidth}}}
		\\
	\subfloat[]{\label{fig:method:edge_grad}{\epsfig{figure=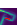,width = 0.18\textwidth}}}
	\quad
	\subfloat[]{\label{fig:method:edge_rga}{\epsfig{figure=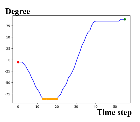,width = 0.2\textwidth}}}
		\\ 
	\subfloat[]{\label{fig:method:transition_hist}{\epsfig{figure=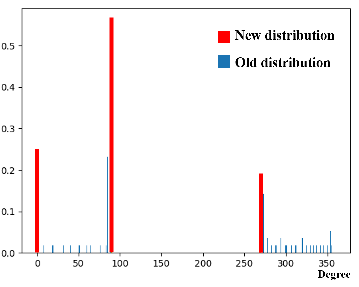,width = 0.23\textwidth}}}
	\,
	\subfloat[]{\label{fig:method:transition_rga}{\epsfig{figure=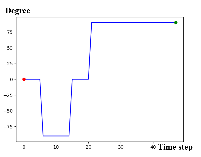,width = 0.23\textwidth}}}

	\caption{a) A transition signal marked in yellow. Red: sequence start. Green: sequence end. Purple: previous structure edge. Brown: next structure edge.
	         b) A candidate edge marked in yellow. Red: its estimated edge line after angle adjustment.
	         c) The candidate edge marked in red on $P'_n$, the first order derivative of probability map $P_n$.
	         d) The candidate edge marked in yellow on the transition angle signal.
	         e) Histogram of the transition angle signal before (blue) and after (red) energy minimization.
	         f) The updated transition angle signal after energy minimization.}
\end{figure}

We adopt the divide and conquer strategy to minimize the energy function $f_S(\Theta)$ . 
Let a contour angle signal between two structure angles be a transition angle signal $\Theta_T$, and let its corresponding contour signal be a transition signal $T$.
We minimize $f_S(\Theta)$ over contour angle signal $\Theta$ by minimizing the same energy function over each transition angle signal $f_S(\Theta_T)$. 
Figure \ref{fig:method:rga1_pts} shows a transition signal and the blue colored line in Figure \ref{fig:method:edge_rga} shows its corresponding transition angle signal.

Let a set of consecutive identical contour angles and their contour points represent a candidate edge.
We want to identify if a candidate edge is noise.
Candidate edges with a structure angle are not noise, and we call them structure edges.
For every structure edge, we fit a line using Least Square Estimation (LSE)~\cite{least_square}, and we call this line an edge line.
These lines will be used to estimate polygon vertices later. 


Given a Boundary Orientation Relation set $S$ with structure angle set $A_S$, and a transition signal $T$ with its contour angle signal $\Theta_T$.
We minimize the objective function $f(\Theta_T)$ by the following three steps.

First, we divide the transition angle signal $\Theta_T$ into two disjoint sets, a candidate angle set $\Theta_{T_c}$, and an edge transition angle set $\Theta_{T_t}$.
Their corresponding contour sets are $T_c$ and $T_t$.
Details are described in Section \ref{sec:transition_classification}.

Second, for each angle in candidate angle set $\Theta_{T_c}$, we quantize it to the nearest structure angle in $A_S$.
We obtain a new contour angle signal $\Theta'_{T_c}$.
As a result, the probability of the candidate angle set containing the relation set $S$ becomes 1:
\begin{equation}
P_{\Theta'_{T_c}}(S) = 1
\end{equation} 
and 
\begin{equation}
f_S(\Theta'_{T_c}) = 0.
\end{equation}
Details are described in Section \ref{sec:method:adj_angle}.

Finally, for contour points in the edge transition contour set $T_t$, we replace them with intersections between edges.
As a result, gradual angle transitions are replaced with dramatic angle change.
Moreover, adding intersections between candidate edges does not change the relative gradient angle of each edge.
We obtain a new edge transition angle set $\Theta'_{T_t}$, and all angles in $\Theta'_{T_t}$ are structure angle.
Now, the probability of edge transition angle set containing the relation set $S$ becomes 1:
\begin{equation}
P_{\Theta'_{T_t}}(S) = 1
\end{equation} 
and 
\begin{equation}
f_S(\Theta'_{T_t}) = 0.
\end{equation}
Details are described in Section \ref{sec:method:transition_analysis}.

Since $\Theta_T$, $\Theta_{T_c}$ and $\Theta_{T_t}$ are sets, and $\Theta_T = \Theta_{T_c} + \Theta_{T_t}$, 
\begin{equation}
f_S(\Theta_T) = f_S(\Theta_{T_c}) + f_S(\Theta_{T_t}) = 0
\end{equation}

\subsection{Noise Removal}
\label{sec:transition_classification}

Candidate edges in transition angle signals do not have a structure angle, because the contour of a network's output may not contain building corners or part of building edges due to noise, smooth filtering, and occlusions.
In these cases, a contour shape may not have enough information to determine whether a candidate edge is a building edge or noise.
Thus, additional information such as texture and edge features of the object needs to be used.

In this paper, we use network's probability map $P_n$ as additional information because CNN considers both texture and edge features.
Ideally, an object's contour aligns with its edge in the probability map.
Let $P'_n$ to be the first order derivative of the probability map obtained by applying Sobel filters \cite{sobel} and normalized to range $[0, 1]$.
Figure \ref{fig:method:edge_grad} shows the first order derivative map $P'_n$ of a building corner.
We obtain an edge map $m_e$ from $P'_n$:
\begin{equation}
m_e(x, y) = \begin{cases}
			1 & P'_n(x, y) > \tau \\
			0 & otherwise,
			\end{cases}
\end{equation}
where $\tau$ is a threshold.
Based on the assumption, any contour point $c$ in a contour signal $C$ should also be an edge point in $m_e$, so $m_e(c) = 1$.
We divide the contour signal $C$ into subsets. 
Each subset contains a consecutive contour signal that describes a candidate edge.
Let $C_e$ be the contour set for the candidate edge $e$.
Due to occlusion and noise, part of $C_e$ may not align with edge responses in $m_e$.
The edge response set $\{ m_e(c) | c \in C_e \}$ of the candidate edge $e$ may contain multiple 0s.  
We assume candidate edge $e$ can be trusted only if at least one point in $C_e$ is an edge point.
If edge $e$ does not include any edge points, we move its contour signal from candidate contour set to edge transition contour set.
We obtain the new candidate contour set $T'_c$ and edge transition contour set $T'_t$:
\begin{equation}
T'_c = \begin{cases}
		T_c & max(\{ m_e(c) | c \in C_e \}) > 0 \\
		T_c - C_e & otherwise,
		\end{cases}
\end{equation}
\begin{equation}
T'_t = \begin{cases}
		T_t & max(\{ m_e(c) | c \in C_e \}) = 0 \\
		T_t + C_e & otherwise.
		\end{cases}
\end{equation}

\subsection{Relative Gradient Angle Quantization}
\label{sec:method:adj_angle}
Let $\Theta_e$ be the contour angle signal of a candidate edge $e$. 
From the candidate edge definition, we have all angles in $\Theta_e$ equal to some angle $\theta_e$.
Let $a_e$ to be $\theta_e$'s nearest structure angle in $A_S$.
We quantize all angles in $\Theta_e$ to $a_e$. 
With the new angle, we fit a line for the candidate edge $e$ using LSE under the constrain that the line slope is $\tan{(a_e+90)}$.
Figure \ref{fig:method:edge_new_line} shows an example of estimated line after angle adjustment.

\subsection{Edge Transition Analysis}
\label{sec:method:transition_analysis}

\begin{figure}[t]
	\centering
	\subfloat[]{\label{fig:method:intersection}{\epsfig{figure=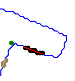,width = 0.18\textwidth}}}
	\qquad
	\subfloat[]{\label{fig:method:parallel}{\epsfig{figure=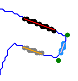,width = 0.18\textwidth}}}
	
	\caption{a) Computing the intersection between two edges. 
	         b) Solving the intersection between two parallel edges (black edge and gray edge) by introducing a new edge (light blue edge).}
\end{figure}

At this point, all the structure edges and trusted candidate edges contain lines describing object's edge.
We aggregate all structure edges and trusted candidate edges into an edge set, and solve the transition between consecutive edges in the set.

If the transition between two edges is non-parallel, we compute the intersection of the two estimated edge lines.
Figure \ref{fig:method:intersection} shows an example of solving non-parallel transition.

If the transition between two edges is parallel, we create a new edge $e_n$ between the two edges to help the transition.
Let $e_1$ and $e_2$ be two consecutive edges that are parallel to each other.
Let $c_i \in e_1$ and $c_j \in e_2$ be the closest pair of points.
We project $c_i$ and $c_j$ onto the line of $e_1$ and obtain $p_1$ and $p_2$.
The horizontal difference is $d_x = x_{p_2} - x_{p_1}$ and vertical difference is $d_y = y_{p_2} - y_{p_1}$.
The middle point is $p_m = (x_{p_1} + \frac{d_x}{2}, y_{p_1} + \frac{d_y}{2})$.  
If there are at least two samples within the middle 50\% vertical range $[y_{p_1} + \frac{1}{4}{d_y}, y_{p_2} - \frac{1}{4}{d_y}]$, we then use these samples to represent the new edge $e_n$.
Otherwise, we use the middle points  
$
C_{e_n} = \{ (x_{p_m}, y_{p_1}), (x_{p_m}, y_{p_m}), (x_{p_m}, y_{p_2}) \}
$
to represent the new edge $e_n$.
The angle of $e_n$ is then quantized and the line of $e_n$ is estimated according to Section \ref{sec:method:adj_angle}.
Intersections for $e_1$ and $e_n$, and $e_2$ and $e_n$ are then computed.
Figure \ref{fig:method:parallel} shows an example of solving parallel transition.

After all the steps, we minimize the objective function $f_s$ to 0, and the set of all intersections between edges represents object's polygon.
Figure \ref{fig:method:rga1_pts} shows a transition signal.
Figure \ref{fig:method:transition_hist} shows the histogram before and after minimizing the objective function.
Figure \ref{fig:method:transition_rga} shows the updated transition angle signal after minimizing the objective function.
Figure \ref{fig:method:example} shows the overall result for the example.

\section{Experimental Results}

\begin{figure*}[t]
	\centering
	\centerline{{\epsfig{figure=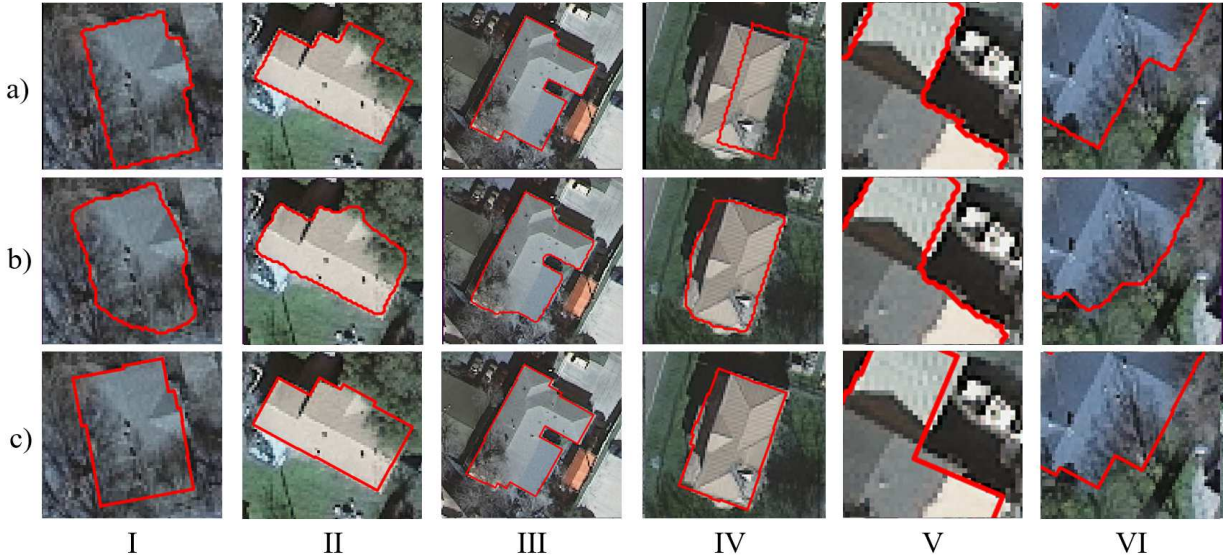,width = 1\textwidth}}}

	\caption{a) Ground truth contours.
	         b) Contours of segmentation mask from network outputs.
	         c) Polygon extracted by our method.
	         I, II show examples of reconstructing building corners and straightening wavy edges.
	         III shows an example of a building with a complex shape.
	         IV shows an example of inaccurate label due to ortho-rectification in the dataset.
	         V shows an example of our method over-straightening the details on a building boundary.
	         VI shows an example of our method reconstructing corners on network's false positive detection.}
	\label{fig:result}
\end{figure*}


In this experiment, we evaluate our method on the Inria Building Dataset~\cite{inria_dataset}.
Particularly, we test our method on buildings with orthogonal corners to demonstrate the feasibility of our method on a set of predefined building corner angles.
Buildings with other types of corners will be tested in future works.
The Inria dataset contains high-resolution orthorectified aerial images with label masks for buildings.
We use image tiles from Austin region for network training and method testing because there are more residential buildings with orthogonal corners in the region.
There are 36 image tiles, and each tile has a size of $5000\times 5000$ pixels, covering an area of 1500 $\times$ 1500 $m^2$ at 30 cm spatial resolution.
We use 8 image tiles for network training, and 28 image tiles for testing.
Our method is evaluated on the buildings that do not touch image border, because image border creates non-orthogonal corners.
The total number of buildings in the test set is 11362. 

For CNN, we use PSPNet \cite{zhao2017pyramid} for demonstration. 
Our method works with any segmentation CNN.
The output probability maps from a PSPNet are used as the inputs of our method.
Because we evaluate our method on buildings with orthogonal corners, we use one Boundary Orientation Relation Set $\{90, 180, 270\}$, and the BORS bank $\phi$ is $\phi = \{\{90, 180, 270\}\}$.
The moving average window size $W_a$ is set to 11 empirically for removing noise and spatial aliasing.
The median filter window size $W_m$ is set to 11 empirically for removing noise in contour angle signals.
Edge threshold $\tau$ is set to 0.1. 
We experimented $\tau$ from 0 to 0.95 with a step size of 0.05, and 0.1 provides the best result.
Higher $\tau$ value represents less detected edge and more shape reconstruction.

From the experiment, PSPNet achieves a 80\% Intersection Over Union (IOU) \cite{fcn}.
Based on the PSPNet outputs, our module converts them into polygons. 
We generate building masks from each polygon, and the masks achieve 78\% IOU.
Although the proposed approach achieves lower IOU than the PSPNet contours, we visually find that our method demonstrates strong capability of reconstructing building corners especially for orthogonal corners, and straightening wavy edges (as shown in Figure \ref{fig:result}), which is essential in most building detection applications where realistic reconstructed building shapes with realistic corners and edges are more important than the mask accuracy that is reflected by IOU.
We attribute the lower IOU of our method to the following facts.
(i) The orthogonal corner assumption in our method is too strong to account for non-orthogonal corners in some buildings.
(ii) Our method is not robust enough to the network's false positive detection that leads to amplification of the network's mistake.
(iii) Ground truth edges sometimes are not accurate as demonstrated in Figure \ref{fig:result} IV. 
In our future research, we will improve our method by (1) using more realistic corner assumptions in addition to the orthogonal corner assumption, (2) preventing over-straightening edges that cause missing building details (as shown in Figure \ref{fig:result} V), and (3) enhancing the robustness of our method to the network's false positive detection to prevent amplification of the network's mistake (as shown in Figure \ref{fig:result} VI).  

\section{Conclusion}
This paper presents a method to extract building polygons from CNN segmentation outputs.
A new transform, Relative Gradient Angle Transform, is described for converting object contour signals into time vs. angle domain.
A new shape descriptor, Boundary Relative Orientation Set, is proposed to represent angle relationship for object's contour.
An energy minimization framework that makes use of the angle relationship in BORS is proposed to straighten edges and reconstruct sharp corners, and the resulting corners create a polygon.
Experimental results demonstrate our method refines CNN output from a rounded approximation to a more clear-cut angular shape of the building footprint.
In the future, we will learn BORS from building labels to handle more building types, such as buildings with parallelogram shapes.
We will investigate adaptive window sizes for smoothing filters to improve our method on detail handling.
We will also investigate incorporating CNN features into our method to address corner reconstructions on false positives.

{\small
\bibliographystyle{ieee_fullname}
\bibliography{ref}
}

\end{document}